\documentclass{article}
\usepackage{ifpdf}
\pdfoutput=1
\usepackage[preprint]{spconf}
\usepackage{amsmath,epsfig}
\usepackage{amsfonts}

\def\deg{{$^\circ$}}

\title{Rollable Latent Space for Azimuth invariant SAR Target Recognition}
\name{Kazutoshi Sagi, Takahiro Toizumi, and Yuzo Senda}
\address{Data Science Research Laboratories, NEC corporation}
                 \toappear{Accepted in {IGARSS2018,
                    July 22-27, 2018, Valencia, Spain}}

\begin{document}
\thispagestyle{empty} 
%
\maketitle
\begin{abstract}
This paper proposes rollable latent space (RLS) for an azimuth invariant synthetic aperture radar (SAR) target recognition.
Scarce labeled data and limited viewing direction are critical issues in SAR target recognition.
The RLS is a designed space in which rolling of latent features corresponds to 3D rotation of an object.
Thus latent features of an arbitrary view can be inferred using those of different views.
This characteristic further enables us to augment data from limited viewing in RLS.
RLS-based classifiers with and without data augmentation and a conventional classifier trained with target front shots are evaluated over untrained target back shots.
Results show that the RLS-based classifier with augmentation improves an accuracy by 30\,\% compared to the conventional classifier.
\end{abstract}
\begin{keywords}
SAR, ATR, autoencoder, latent feature, view, disentangle
\end{keywords}
\section{Introduction}
\label{sec:intro}

Automatic target recognition (ATR) on synthetic aperture radar (SAR) imagery is under keen interest especially in social security and defense applications.
SAR ATR was considered difficult since human operators are no good at interpreting SAR images and crafting good manual features for ATR \cite{saratrsurvey16}.
After deep neural network (DNN) techniques are introduced, SAR ATR reaches a feasible level  \cite{dlsurvey17}.

Several classification approaches have already been proposed in SAR ATR.
Wilmanski et al. applied convolutional neural network (CNN) to SAR classification problem and showed promising results of high accuracy \cite{wilmanski2016modern}.
The method achieved 97\,\% in accuracy while manual feature based classifiers showed approximately 70\,\%.
The above paper relies on Moving and Stationary Target Acquisition and Recognition (MSTAR) dataset  \cite{ross1998standard}.
The dataset contains SAR images of 360 viewing directions of 10 military ground target vehicles from depression angles of 15\deg\ and 17\deg.
Usually all SAR images from depression angle of 17\deg\ were used to train classifiers, but that is unrealistic in actual applications.
There are still challenges due to limited viewing direction and scarce labeled data.

To obtain intrinsic features invariant to viewing direction, Song et al. used a generative DNN framework to capture the features and the viewing direction of a target \cite{song17}.
The framework consists of a constructor, a generator, and an interpreter.
The constructor uses labels of training data to form a continuous target feature space independent of viewing direction.
The generator and the interpreter learn correspondences between SAR images and sets of point in the feature space and viewing direction.
Even though the feature space is continuous, its representation capability could be limited to variety of the labels.

To overcome the challenges in a different way, Avolio et al. proposed use of a SAR simulator to generate plenty of training data from computer-aided design (CAD) models \cite{Avolio17}.
A classifier can be trained with the generated data, but CAD models are necessary instead.

Several approaches have been proposed for 3D view synthesis in other computer vision applications such as robotics, rendering and modeling.
One of them is to employ conditional autoencoder to disentangle latent vactors into two parts, i.e. pose and identity \cite{yang2015weakly}. 
The autoencoder modifies the pose part to produce a different view of an object.
The identity part contains individuality of an object and thus it can be used for recognition.

A purpose of this paper is to propose new disentangling using a novel autoencoder.
The novel autoencoder network is developed for learning 3D rotations of objects in SAR images.
The learned autoencoder provides a discriminative view-invariant feature space.
This disentangling can handle any SAR images without labels or CAD models of targets.

\section{Proposed method}
\label{sec:proposal}
\subsection{Rollable Latent Space}
\label{ssec:rls}
We propose rollable latent space (RLS) to cope with variation from viewing directions.
In object recognition, a latent vector is expected to represent disentangled information of a certain object.
In the RLS, a latent vector consists of multiple sub-vectors, each of which has a fixed length of structural and directional information.
Number of sub-vectors corresponds to the number of equally spaced viewing directions.
The sub-vector can be rolled to represent a 3D rotated image of its structural information.

An RLS encoder converts an image $\mathbf{X}$ seen from a direction $\theta_{i}$ to a latent vector $\mathbf{Z}$,
\begin{equation}
	\mathbf{Z}(\theta_{i}) = \mathit{Encoder}(\mathbf{X}(\theta_{i})).
	\label{eq:zthetai}
\end{equation}
$\mathbf{Z}$ from another direction $\theta_{j}$ can be approximated by simply applying roll function to $\mathbf{Z}$,
\begin{align}
     \mathbf{Z}(\theta_{j}) &= \mathit{Encoder}(\mathbf{X}(\theta_{j})) \\
                                    &\simeq \mathit{Roll}(\mathbf{Z}(\theta_{i}), j-i),
	\label{eq:zthetaj}
\end{align}
where $\mathit{Roll}(\mathbf{Z}, s)$ rolls all sub-vectors of $\mathbf{Z}$ by shift parameter $s$.
In a simple RLS, $\mathit{Roll}(\mathbf{Z})$ can be written as
\begin{equation}
	 \mathit{Roll}(\mathbf{Z}_{k},s) = \mathbf{R}^{s} \cdot \mathbf{Z}_{k},
	\label{eq:roll}
\end{equation}
where $\mathbf{Z}_{k}$ is a sub-vector of $\mathbf{Z}$, and $\mathbf{R}$ is a cyclic permutation matrix.
Roll function can be interpolative to support continuous rotation.

To let  $\mathit{Encoder}$ learn this disentanglement, an autoencoder approach is employed.
An RLS autoencoder is defined as
\begin{equation}
	\mathbf{X}(\theta_{j}) \simeq \mathit{Decoder}(\mathit{Roll}(\mathit{Encoder}(\mathbf{X}_{i}), j-i)).
\end{equation}
By feeding images of known objects to the autoencoder with varying $s$, $\mathit{Encoder}$ and $\mathit{Decoder}$ are obtained.

\subsection{Feature Augmentation in RLS}
\label{ssec:augmentation}
To train a classifier, data augmentation is necessary especially for scarce labeled images like SAR.
In addition to ordinary augmentation techniques, feature-level augmentation can be applied easily in RLS.
A simple rolling function can produce a number of augmented feature vectors when using a latent vector $\mathbf{Z}$ in RLS as a feature vector.
$\mathit{Classifier}$ is trained to fulfill
\begin{equation}
	\mathbf{Y}_{i} = \mathit{Classifier}(\mathit{Roll}(\mathit{Encoder}(\mathbf{X}_{i}), s_{rand})),
\end{equation}
where $\mathbf{Y}_i$ is the label of $\mathbf{X}_i$, when $s_{rand}$ is randomly given for feature augmentation.

\subsection{Implementation}
\label{ssec:implementation}
\begin{figure}[t]
\begin{minipage}[b]{1.0\linewidth}
  \centering
  \centerline{\epsfig{figure=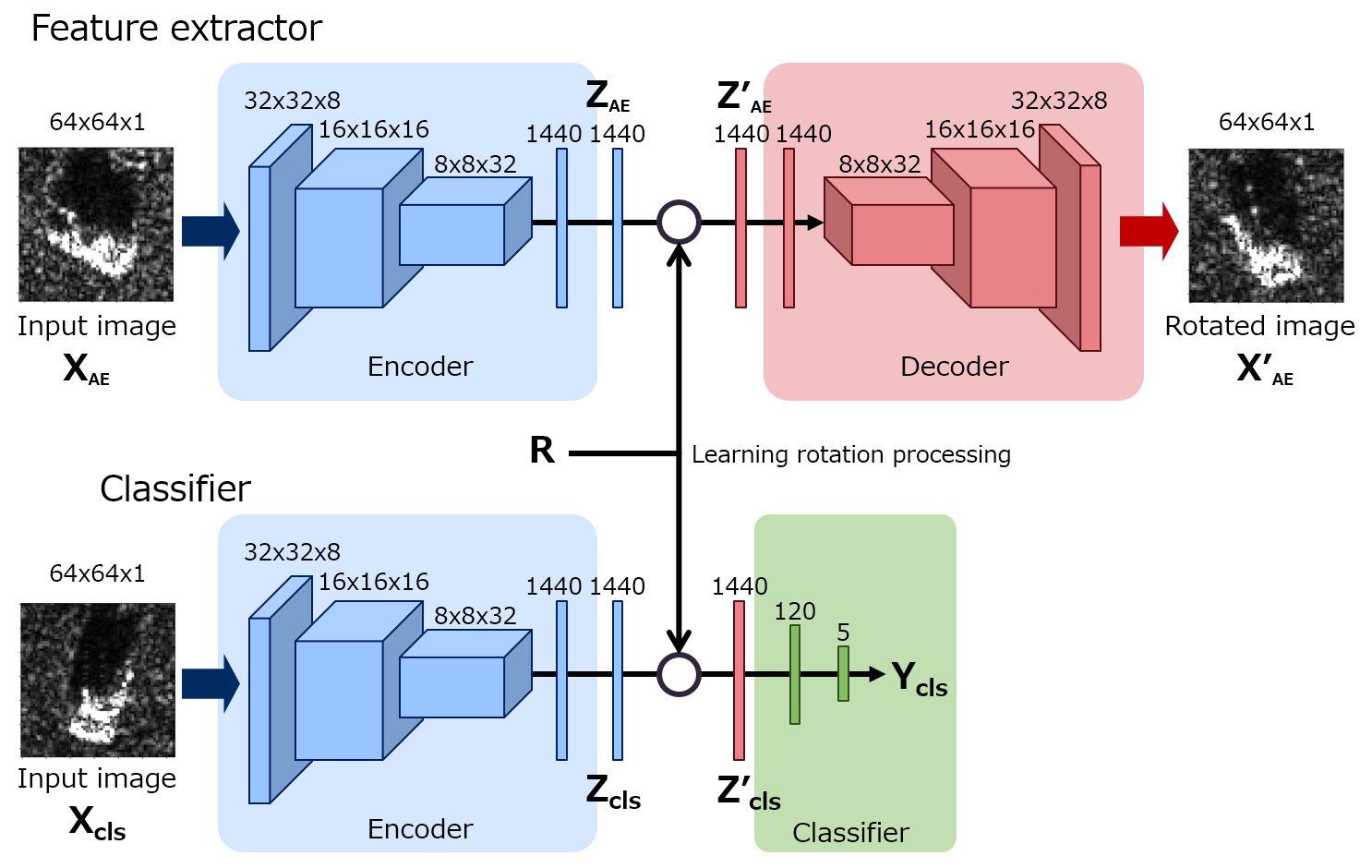,width=9.cm}}
\end{minipage}
\caption{Proposed network architecture for learning a RLS-based classifier.}
\label{fig:network}
\end{figure}

Figure \ref{fig:network} shows an entire structure of RLS-based autoencoder and classifier used in this paper.
To promote disentanglement, variational autoencoder (VAE) is employed in conjunction with CNN \cite{kingma2013auto}.
In this paper a deep convolutional encoder-decoder architecture proposed by Yang et al.\,\cite{yang2015weakly} is applied to the encoder and the decoder.
The encoder network consists of three 5$\times$5 convolution layers with stride 2 and 2-pixel padding and ReLU activation, and two fully connected layers.
Note that employing CNN architecture reduces an error of the target shift of few pixels because of its convolutions and strides.
The decoder takes symmetric architecture to the encoder. 
Fixed stride-2 convolution and upsampling are employed instead of max-pooling and unpooling.
The classifier recieves the feature vector $\mathbf{Z}$ as an input.
The classifier structure employed here simply consists of two additional fully connected layers with nodes of 120 and 5, respectively.

\section{Experiment}
\label{sec:experiment}
\subsection{Setup}
\label{ssec:setup}
For the sake of demonstrating RLS potential, we perform an experiment based on realistic SAR ATR operation.
The experiment procedure is following below,
\begin{enumerate}
\item RLS training using omnidirectional SAR images,
\item classifier training using SAR target front shots,
\item classifier testing for SAR target back shots.
\end{enumerate}

\begin{figure}[t]
\begin{minipage}[b]{1.0\linewidth}
  \centering
  \centerline{\epsfig{figure=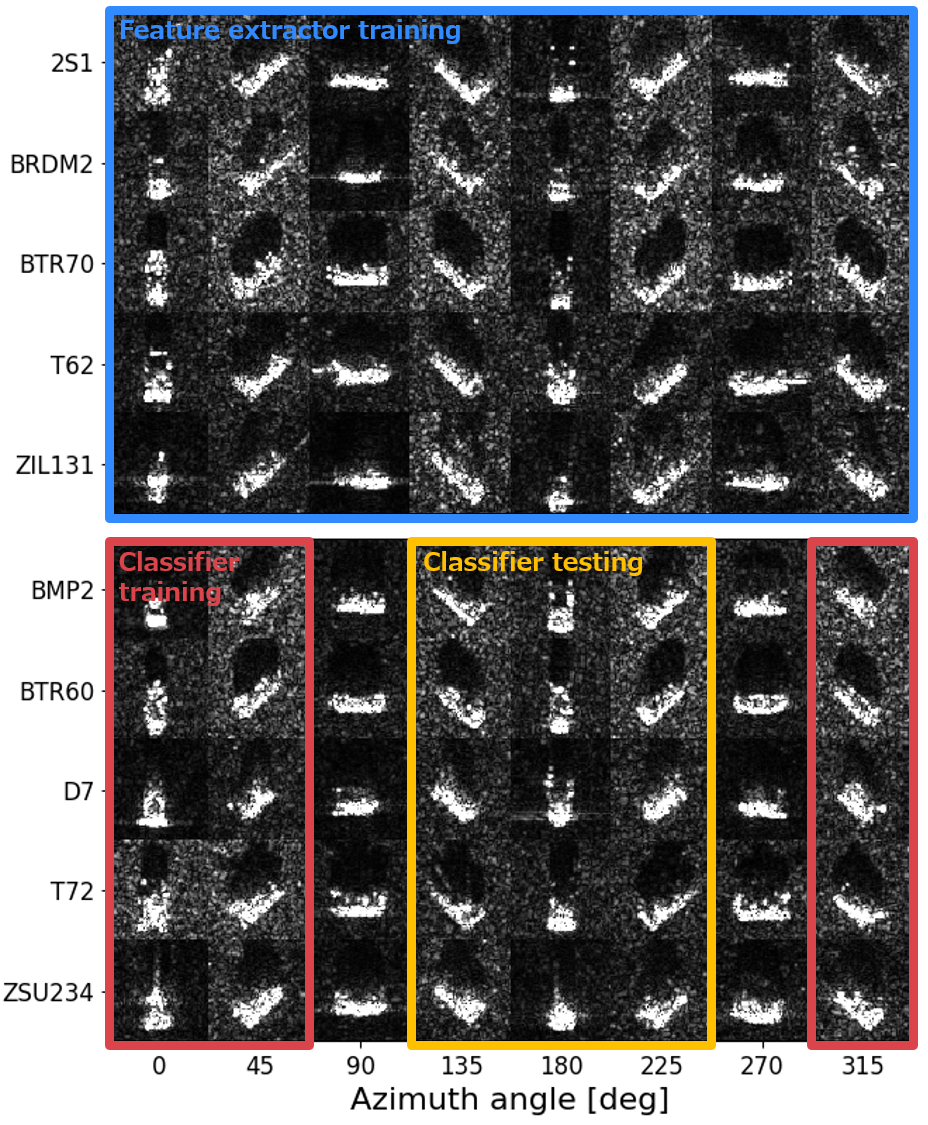,width=8.cm}}
\end{minipage}
\caption{MSTAR SAR sample images of depression angle of 17\deg. Images taken from eight different azimuth angles are shown. Five vehicle classes shown in the top panel (in the blue box) are used to train our auto-encoder. The other five vehicles are used in classification training (in the red box) and testing (in the yellow box).}
\label{fig:mstar}
\end{figure}

MSTAR SAR images of depression angle 17\deg\ are used in this paper.
The 17\deg\ depression angle dataset is separated into three sub groups.
Figure \ref{fig:mstar} presents examples of MSTAR SAR images used in this paper.
All SAR chips are cropped into 64$\times$64 pixels from 128$\times$128 pixels in order to exclude a contribution of background scattering to classification performance. 

For RLS training phase, five vehicle classes are used.
These five vehicle classes are selected to have variation in vehicle type in order to let RLS be generalized.
Each class contains SAR images observed from entire azimuth angles.
Table \ref{tab:mstardata1} describes the detail of selected classes.

\begin{table}[b]
\begin{center}
\caption{Data set used to train Encoder and RLS}
\label{tab:mstardata1}
  \begin{tabular}{ccc}
    \hline\hline
    Class & Azimuth coverage & Size \\
    \hline\hline
    2S1 & 0\deg $\sim$ 360\deg & 299 \\
    BRDM2 & 0\deg $\sim$ 360\deg & 298 \\
    BTR70 & 0\deg $\sim$ 360\deg & 233 \\
    T62& 0\deg $\sim$ 360\deg & 299 \\
    ZIL131 & 0\deg $\sim$ 360\deg  & 299\\
    Total &$$--&1428 \\
    \hline
  \end{tabular}
\end{center}
\end{table}

\begin{table}[!b]
\begin{center}
\caption{Data set used to classifier training and testing}
\label{tab:mstardata2}
  \begin{tabular}{cccc}
    \hline\hline
    &Class & Azimuth coverage & Size \\
    \hline\hline
    &BMP2-C21& $-$45\deg(315\deg) $\sim$ 45\deg & 59 \\
    &BTR60& $-$45\deg(315\deg) $\sim$ 45\deg & 66 \\
    Train&D7& $-$45\deg(315\deg) $\sim$ 45\deg & 77 \\
    &T72-SN132& $-$45\deg(315\deg) $\sim$ 45\deg & 59 \\
    &ZSU234& $-$45\deg(315\deg) $\sim$ 45\deg & 78 \\
    &Total &$-$&339 \\
    \hline
    &BMP2-C21& 135\deg $\sim$ 225\deg & 58 \\
    &BTR60& 135\deg $\sim$ 225\deg & 57 \\
    Test&D7& 135\deg $\sim$ 225\deg & 71 \\
    &T72-SN132 & 135\deg $\sim$ 225\deg & 57 \\
    &ZSU234& 135\deg $\sim$ 225\deg & 78 \\
    &Total &$-$&321 \\
    \hline  \end{tabular}
\end{center}
\end{table}
The other five vehicle classes are used for a performance evaluation of RLS in classification with and without data augmentation in RLS.
\begin{figure}[tb]
\begin{minipage}[b]{1.0\linewidth}
  \centering
  \centerline{\epsfig{figure=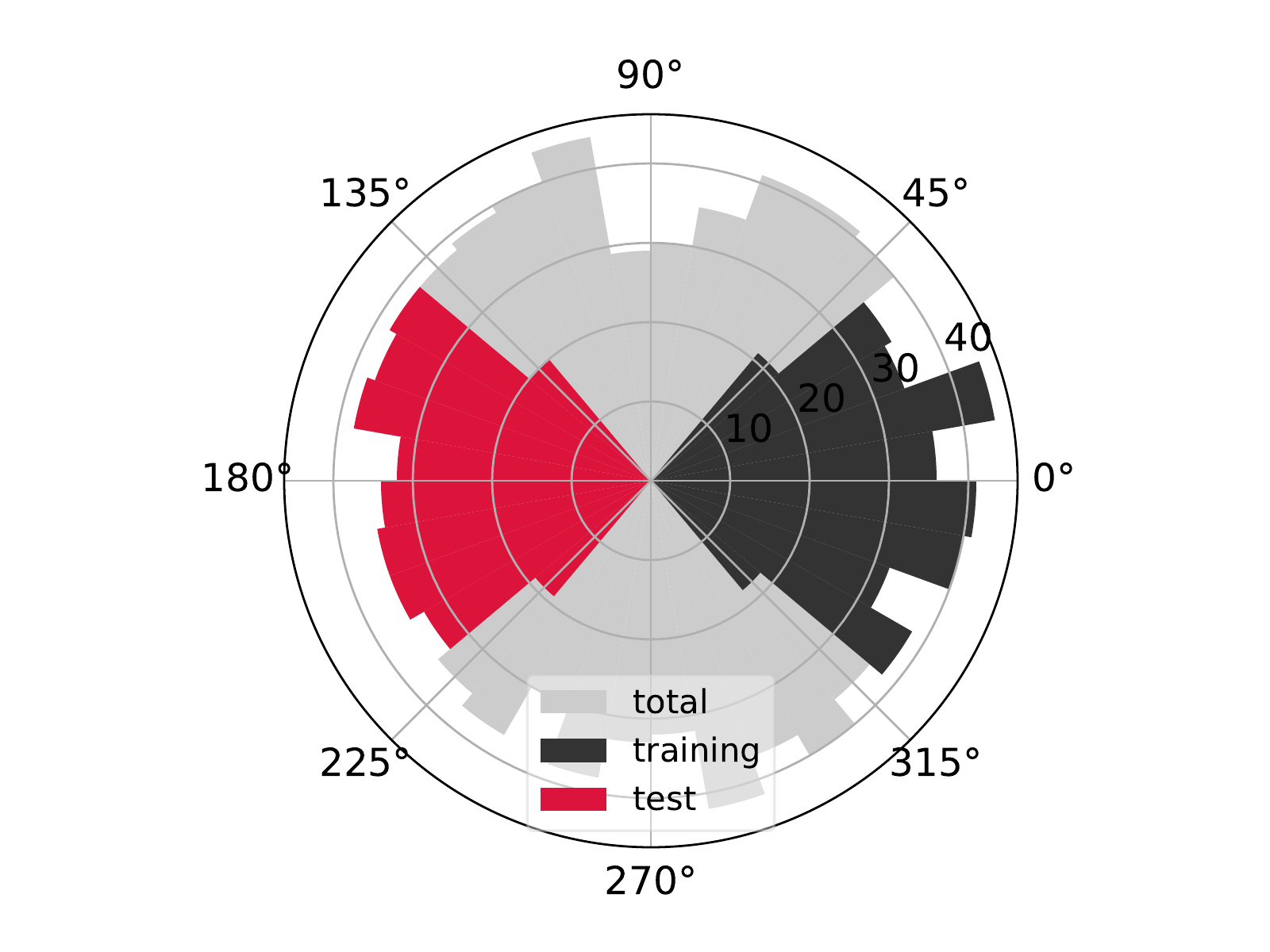,width=8.cm}}
\end{minipage}
\caption{Azimuthal distribution of dataset used in training (dark gray area) and testing (red area) classifier. Light gray area indicates a distribution of all five vehicle images listed in Tab. \ref{tab:mstardata2}. }
\label{fig:azdata}
\end{figure}
Those five vehicles are further separated into two groups by their azimuth angles for classifier training and testing.
Images within an azimuth angle range between $-$45\deg\ and 45\deg\ are used in classifier training. 
Images within an azimuth angle range between 135\deg\ and 225\deg\ are used in classifier testing. 
Table \ref{tab:mstardata2} summarizes the details of data sets used in classifier training and testing.
Figure \ref{fig:azdata} illustrates the azimuthal distribution of training and testing dataset.

\begin{figure}[tb]
\begin{minipage}[b]{1.0\linewidth}
  \centering
  \centerline{\epsfig{figure=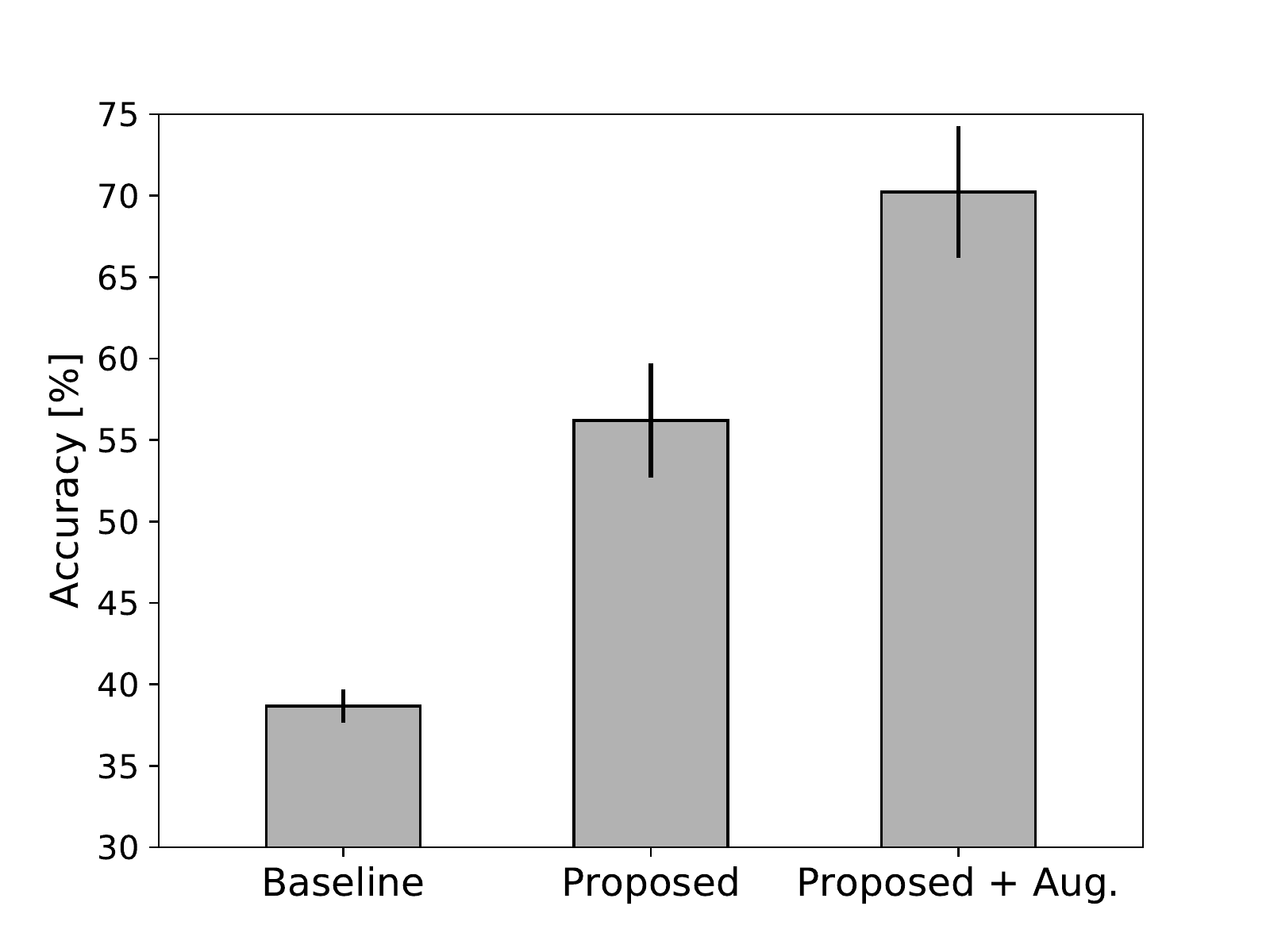,width=8.cm}}
\end{minipage}
\caption{Mean accuracy for MSTAR five class classification over 20 calculations and its $2\sigma$ value as an error. The baseline CNN (left), the proposed methods w/o (middle) and w/ (right) data augmentation in RLS are presented here.}
\label{fig:accs}
\end{figure}

\subsection{Baseline}
\label{ssec:baseline}
A CNN architecture modified from Wilmanski et al.\,\cite{wilmanski2016modern} is selected as a baseline to compare with the proposed method.
The differences between the baseline and the original Wilmanski's network are a size of input image and kernel sizes of three convolutional layers.
The original Wilmanski'{}s CNN uses 128$\times$128 MSTAR chips as input images. 
We modified convolutional layers to fit with the input size of 64$\times$64 pixels.

\subsection{Results}
\label{ssec:result}

Figure \ref{fig:accs} shows classification accuracies of proposed approaches with and without data augmentation in RLS.
The accuracies are averaged over 20 calculations with different initial weights.
Errors are given as 2 standard deviations ($2\sigma$).
Figure \ref{fig:accs} also plots an accuracy of the baseline CNN-based classifier.
The proposed RLS-based classifier with data augmentation achieves the highest accuracy of 70.2\,\% ($\pm$4.04\,\%).
The second highest accuracy of 56.2\,\% ($\pm$3.51\,\%) is for the proposed RLS-based classifier without data augmentation, while the accuracy of the baseline is 38.7\,\% ($\pm$1.03\,\%).
The proposed method with augmentation outperforms the others.
This is because that RLS has successfully obtained an azimuth rotation invariant feature space.

\begin{figure}[tb]
\begin{minipage}[b]{1.0\linewidth}
  \centering
  \centerline{\epsfig{figure=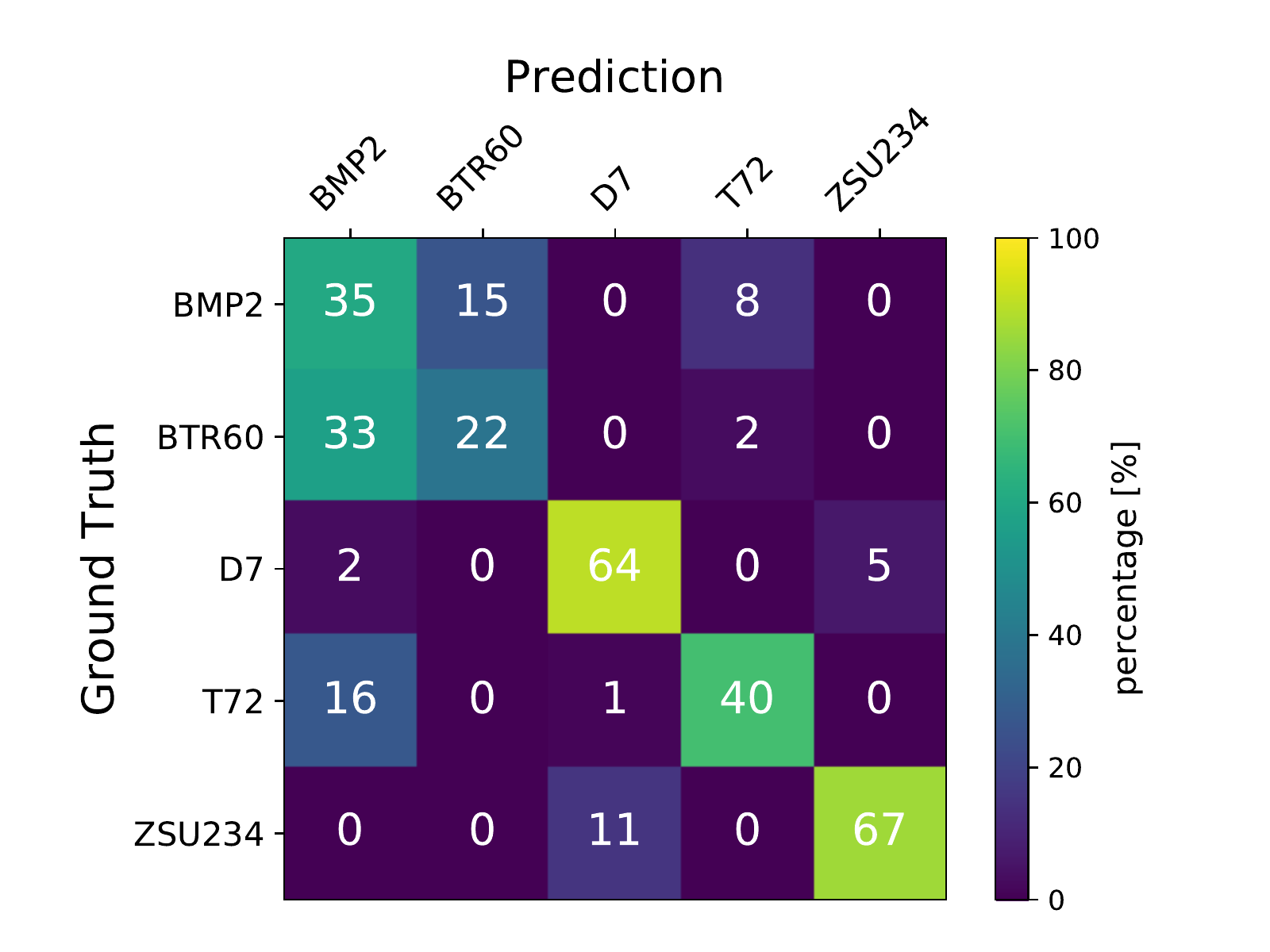,width=8.cm}}
\end{minipage}
\caption{A confusion matrix of the classification result obtained by the proposed method with data augmentation in RLS. Color shows a percentage value.}
\label{fig:cmat}
\end{figure}

Figure \ref{fig:cmat} is a confusion matrix of the classification result obtained by the proposed method with augmentation.
The most confused class is BMP2. 
Especially BMP2 and BTR60 misrecognize each other.
This is due to similarity in $\mathbf{Z}$, which shows there is still scope for improvements in learning RLS.


\section{Conclusions}
\label{sec:conclusion}
This paper has proposed RLS for an azimuth invariant target recognition in SAR images using an autoencoder architecture and a cyclic permutation matrix.
In RLS, rolling of latent features corresponds to 3D rotation of an object.
Latent features of an arbitrary view can be approximated by changing the cyclic permutation matrices, which enables us to augment data from limited viewing.
Evaluations have been performed using MSTAR images.
In the evaluations, we trained an RLS-based classifiers with and without augmentation and a conventional classifier with target front shots, then compared thier accuracies over untrained target back shots.
It is confirmed that the RLS-based classifiers outperformed the conventional classifier.
The RLS-based classifier with augmentation achieved the highest accuracy of 70.2\,\%, which improves an accuracy by 30\,\% compared to the conventional classifier.
\bibliographystyle{IEEEbib}
\bibliography{ref.bib}

\end{document}